\documentclass{l4dc2027}
\usepackage{acronym}
\usepackage{siunitx}
\usepackage{bm}
\usepackage{booktabs}
\usepackage{mathtools}
\usepackage{threeparttable}

\acrodef{spd}[SPD]{symmetric positive definite}
\acrodef{mp}[MP]{motion primitive}
\acrodef{mps}[MPs]{motion primitives}
\acrodef{ds}[DS]{dynamical system}
\acrodef{fk}[FK]{forward kinematics}
\acrodef{llc}[LLC]{low level controller}
\acrodef{cbf}[CBF]{control barrier function}
\acrodef{dof}[DoF]{degrees of freedom}
\acrodef{qp}[QP]{quadratic program}
\acrodef{tamp}[TAMP]{task and motion planning}
\acrodef{lfd}[LfD]{Learning from Demonstration}
\acrodef{dnn}[DNN]{Deep Neural Network}
\acrodef{dsmp}[DS-MP]{DS-MP}
\acrodef{dmp}[DMP]{Dynamic Movement Primitive}
\acrodef{cdmp}[ONCol-DMP]{Optimized Normalized Collaborative DMP}
\acrodef{dsmpman}[CDSM]{Curve-induced Dynamical systems on Smooth Manifolds}
\acrodef{puma}[PUMA]{Policy via neUral Metric leArning}
\acrodef{calds}[CALDS]{Cooperate Adaption for Learned Dynamical Systems}

\DeclareSIUnit{\rad}{rad}

\title[Cooperative Mobile Manipulation using Learned DSs]{Cooperative Mobile Manipulation using Learned Dynamical Systems}
\usepackage{times}

\author{%
 \Name{Saray Bakker} \Email{s.bakker-7@tudelft.nl}\\
 \addr Faculty of Mechanical Engineering, TU Delft, The Netherlands
 \AND
 \Name{Wendelin B\"ohmer} \Email{j.w.bohmer@tudelft.nl}\\
 \addr Faculty of Electrical Engineering, Mathematics \& Computer Science, TU Delft, The Netherlands
  \AND
 \Name{Javier Alonso-Mora} \Email{j.alonsomora@tudelft.nl}\\
 \addr Faculty of Mechanical Engineering, TU Delft, The Netherlands
}
\makeatletter
\def\ps@jmlrtps{%
  \let\@mkboth\@gobbletwo
  \def\@oddhead{}%
  \def\@evenhead{}%
  \def\@oddfoot{\hfil\thepage\hfil}%
  \def\@evenfoot{\hfil\thepage\hfil}%
}
\makeatother
\begin{document}
\maketitle

\begin{abstract}%
Cooperative mobile manipulation requires robots to coordinate their motions while ensuring whole-body collision avoidance and preserving task-relevant behaviors.
We propose CALDS (Cooperative Adaptation of Learned Dynamical Systems), a framework that enables mobile manipulators to adapt learned single-agent behaviors for coordinated execution. 
Building on individual dynamical systems (DSs) learned from demonstrations, CALDS enables cooperative adaptation through two key components: Cooperative Policy Selection, which identifies compatible combinations of learned policies, and Temporal Adaptation, which ensures collision-free cooperation while preserving the spatial structure and convergence properties of the original behaviors.
We evaluate CALDS across three diverse scenarios: assistive dressing, restaurant cleaning and serving, and drawer opening and object placement.
Quantitative results demonstrate its effectiveness in improving task success and enabling safe, coordinated simultaneous execution while maintaining fidelity to demonstrated behaviors.
\end{abstract}


\section{Introduction}
Mobile manipulators are versatile platforms capable of assisting humans in household tasks~\citep{wu2023tidybot}, and performing labor-intensive jobs, such as operating in greenhouses~\citep{sanchez2024greenhouses} or construction sites~\citep{hartmann2022long}. In these domestic and professional environments, efficiency rises when robots collaborate with other agents, anticipating on their actions and intentions. 
For instance, consider a robot assisting a human in putting on a jacket while another shows how the outfit looks using a mirror, or a restaurant where one robot places a glass on the table-area cleaned by another robot.
Such scenarios require robots to proactively adapt their behaviors to other agents, ensuring both task effectiveness and safety.

\ac{lfd} has emerged as a powerful framework for transferring human skills to robots, with successful applications in manipulators~\citep{reuss2023goal, perez2024puma}, dual-arm systems~\citep{zhao2023learning}, and mobile manipulators~\citep{fu2024mobile, sundaresan2025homer, dong2025m4diffuser}. However, extending these skills to collaborative behaviors involving \emph{multiple mobile manipulators} presents several challenges.
Firstly, collecting multi-agent demonstrations is inherently complex, making even a small number of demonstrations difficult to obtain. 
Secondly, mobile manipulators operate in cluttered and dynamic environments, requiring whole-body collision avoidance with other robots and surrounding obstacles. Existing approaches often delegate collision avoidance to low-level planners~\citep{sundaresan2025homer}, whose deviations from demonstrated motions can compromise task execution and push learned policies into off-distribution states, resulting in potentially unreliable or unintuitive actions.
These challenges motivate frameworks that enable collaborative mobile manipulation from small amounts of demonstrations, while accounting for whole-body constraints and anticipating the actions of other agents.

\begin{figure}
    \centering
    \includegraphics[width=\linewidth]{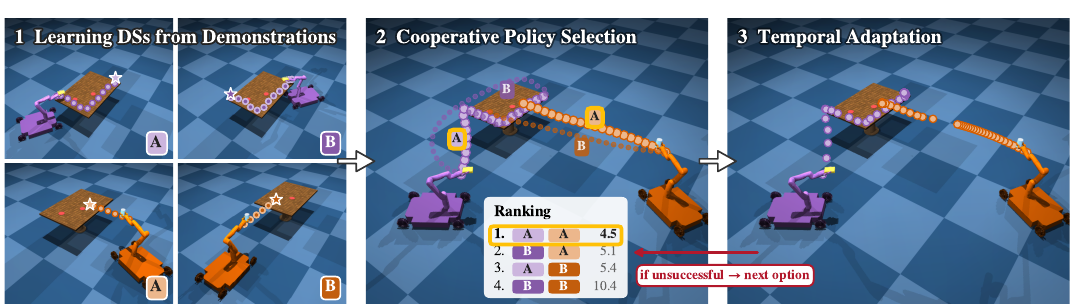}
    \caption{
    Overview of \ac{calds} for cooperative multi-agent adaptation in a restaurant scenario, where one robot cleans a table while the other places a cup on the cleaned area.
    Firstly, we generate a set of learned single-agent \acp{ds} for each task. In the multi-agent context, Cooperative Policy Selection ranks and selects the most promising combination, which Temporal Adaptation then adjusts to satisfy task constraints while ensuring whole-body collision-free trajectories, selecting the next-best combination if unsuccessful. 
}
    \label{fig:overview}
\end{figure} 
Rather than depending on extensive multi-agent demonstrations, recent approaches have explored composing multi-agent behavior from single-agent demonstrations. 
\cite{singh2025collaborative} define \acp{cdmp} to generate collaborative trajectories leveraging a time-heuristic and end-effector spatial collision avoidance to modulate learned skills. 
Furthermore, single-agent demonstrations can be combined with multi-agent cost functions to train coordinated dual-manipulator diffusion policies~\citep{peters2026coordinated}. Extending this approach to mobile manipulators, however, requires reliable visual perception for both diffusion-based policy generation and whole-body collision avoidance.
Other approaches learn collaborative behaviors from multi-manipulator demonstrations, while focusing on maintaining global consistency during decentralized execution~\citep{he2025latent, dong2025mimic}.
Learning-free task and motion planning (TAMP) methods, such as those using temporal logic~\citep{saha2017task} or long-horizon planning frameworks~\citep{hartmann2022long}, also offer solutions for collaborative tasks incorporating whole-body collision avoidance. However, their computational demands can limit real-time applicability. Moreover, collision-free collaboration does not necessarily imply efficient execution: sequential motion or waiting for other robots to pass can cause unnecessary delays, while concurrent execution requires explicit coordination to avoid collisions~\citep{rudy2022multiple}.

To address these challenges, we propose \ac{calds} (\emph{Cooperative Adaptation of Learned Dynamical Systems}), a framework that enables mobile manipulators to coordinate learned single-agent behaviors in shared workspaces while ensuring whole-body collision avoidance and task constraints (Fig.~\ref{fig:overview}).
Each robot is equipped with a set of learned \acfp{ds}, which encode diverse strategies for accomplishing the assigned task. These learned \acp{ds}, such as proposed in~\citep{urain2022learning, perez2024puma, bakker2026curve}, can be generated from only a few (e.g. 1--10) demonstrated end-effector pose trajectories in $SE(3)$, while guaranteeing convergence to the desired goal. 
\ac{calds} comprises two centralized stages: \emph{Cooperative Policy Selection}, which identifies compatible combinations of single-agent policies, and \emph{Temporal Adaptation}, which adjusts their temporal profiles to achieve collision-free collaboration while preserving the spatial structure, interpretability, and convergence of the original learned \acp{ds}. 
Both stages generate trajectories in which the adapted learned DSs drive the motion at every step, leveraging JAX~\citep{jax2018github, jaxopt_implicit_diff} and FRAX~\citep{morton2026frax} kinematics for batched, differentiable computation.
We validate the proposed framework through quantitative experiments in three collaborative scenarios. We compare \ac{calds} against existing collaborative policy adaptation~\citep{singh2025collaborative}, whole-body collision avoidance via \acp{cbf}~\citep{singletary2021safety} and nominal single-agent policies, demonstrating its effectiveness in enabling safe and task-consistent operation.
\section{Preliminaries on Learned Dynamical Systems}
\acp{ds} provide an effective framework for encoding motion behaviors either from human demonstrations~\citep{billard2022learning} or through manual design by combining multiple sub-behaviors~\citep{ratliff_riemannian_2018}. A motion is represented as an autonomous continuous-time \acs{ds} of the form,
\begin{equation} \label{eq:1_DS}
    \dot{\bm{x}}(t) = f(\bm{x}(t)),
\end{equation}
where $\bm{x}(t)$ denotes the system state and $\bm{x}^{\ast}$ is the equilibrium point corresponding to the task goal, such that $f(\bm{x}^{\ast}) = \bm{0}$, with $\bm{0}$ denoting the zero vector of appropriate dimension.
The learned behavior is typically defined in the end-effector state space, which enables adaptation to changes in goal pose and the robot's embodiment. In this work, we represent the end-effector pose as $\bm{x} = [\bm{p}, \bm{\xi}]$, where $\bm{p}$ denotes the Cartesian position and $\bm{\xi}$ denotes the orientation represented as a quaternion. The output of the \acs{ds}, $\dot{\bm{x}}$, therefore corresponds to the end-effector twist.

Stability of learned \acp{ds} can be enforced in several ways. In \citep{urain2022learning}, stability is achieved through a diffeomorphic mapping between a simple stable system and task-space dynamics, extending the formulation to Lie groups. Other approaches enforce stability through the training objective of \acp{dnn}~\citep{perez2024puma}, or guarantee convergence toward a geometric curve learned from demonstrations \citep{bakker2026curve}. More recently, \citep{perez2026letthedynamicsflow} incorporated Lyapunov stability constraints into flow-matching-based \acp{ds}, including Lie group formulations.
\section{Problem formulation} \label{sec: problem formulation}
Consider a multi-robot system consisting of $N$ mobile manipulators. Each robot $i \in [N]$, using shorthand notation $[N] = \{1,\ldots,N\}$, is assigned a task $\mathcal{T}_i$, which can be executed using one of $M_i$ distinct learned \acp{ds}.
\begin{equation} \label{eq: 2_DS}
\dot{\bm{x}}_i = \bm{f}_{i,m_i}(\bm{x}_i),
\quad m_i \in [M_i].
\end{equation}
We refer to these learned \acp{ds} as \emph{task strategies}, each encoding a single-agent strategy for the assigned task without considering interactions with other agents.

Our objective is to select an appropriate combination of task strategies and adapt their execution to enable safe and effective collaboration, while minimizing deviations from the original learned behaviors. Specifically, we formulate the following optimization problem:
\begin{subequations} \label{eq: problem_optimization}
\begin{align}
\underset{
\substack{
m_i\in[M_i],\\
\bm{v}_i(t)
}}
{\operatorname{minimize}}
\quad &
\sum_{i=1}^{N} \left(
\int_0^T
d\left(\bm{v}_{i}(t), \bm{f}_{i,m_i}(\bm{x}_i(t))\right)^2 dt + J(m_i) \right)
\\
\text{s.t.} \quad &
\dot{\bm{q}}_i(t) = \bm{h}(\bm{q}_i(t), \bm{v}_i(t)), \ \ \ \ \ \hspace{0.5pt} \quad \forall i\in[N] && \triangleright\ \text{Robot dynamics} \label{robot_dynamics_problem_def}\\
&
\mathcal{V}_i(\bm{q}_i(t))
\cap
\mathcal{V}_j(\bm{q}_j(t))
=\emptyset, 
\quad \forall i\neq j
&& \triangleright\ \text{Collision avoidance} \label{eq: coll_avoid_problem_def}
\\
&
t_i^{\text{c}}\leq t_j^{\text{c}},
\quad \ \ \ \ \ \   \ \ \ \ \ \   \ \ \ \ \ \   \ \ \ \ \ \    \ \ \ \ \ \ \hspace{2.0pt} \forall(i,j)\in\mathcal{P}
&& \triangleright\ \text{Task ordering} \label{eq: task_order_problem_def}
\\
&
\lim_{t \rightarrow \infty} d(\bm{x}_i(t), \bm{x}_{i}^{\text{g}}) = 0,
\quad \ \ \ \ \ \  \hspace{1pt} \forall i\in[N]
&& \triangleright\ \text{Convergence} \label{eq: convergence_problem_def}
\end{align}
\end{subequations}
Here, $\bm{v}_i(t)$ denotes the adapted velocity action, $d(\cdot, \cdot)$ an appropriate distance metric, $J(m_i)$ a task-strategy dependent cost, and~\eqref{robot_dynamics_problem_def} the robot dynamics. 
Constraint~\eqref{eq: coll_avoid_problem_def} enforces whole-body collision avoidance between the robots throughout the horizon, where $\mathcal{V}_i(\bm{q}_i)$ denotes the volume occupied by robot $i$ at configuration $\bm{q}_i$. 
Constraint~\eqref{eq: task_order_problem_def} accounts for temporal dependencies between tasks, where $\mathcal{P}$ denotes the set of robot pairs subject to ordering constraints, and $t_i^{\text{c}}$ represents the completion time of task $\mathcal{T}_i$. Finally, constraint~\eqref{eq: convergence_problem_def} ensures that the adapted execution retains the convergence properties of the selected learned \acp{ds}, such that each robot ultimately reaches its corresponding goal state.

\section{\acl{calds}}
In this section, we introduce \ac{calds} (Fig.~\ref{fig:overview}). First, we learn distinct \acp{ds} from a small set of trajectory demonstrations to obtain single-agent policies. To enable cooperative multi-agent adaptation, \ac{calds} comprises two layers: Cooperative Policy Selection (Sec.~\ref{sec: policy_selection}) and Temporal Adaptation (Sec.~\ref{sec: temporal_optimization}). Both layers rely on generating robot trajectories along a horizon, as described in Sec.~\ref{sec: trajectory_generation}.

\subsection{Trajectory Generation} \label{sec: trajectory_generation}
Both the Cooperative Policy Selection and Temporal Adaptation require a trajectory and trajectory cost that simulates and evaluates the future actions of the robots. 
We therefore forward-simulate whole-body robot motion over a finite horizon of duration $T$, discretized into $K$ equally spaced intervals of length $\Delta t = T/K$, with time points indexed by $k \in \{0,\ldots,K\}$.

As the learned \acp{ds} generate end-effector velocity commands and take the current end-effector pose as input, we obtain the latter through \ac{fk} from the joint configuration $\bm{q}_i$ of robot $i$, $\bm{x}_i(k)=\mathrm{FK}(\bm{q}_i(k))$. We introduce a velocity scaling factor $\alpha_i(k)$ for each robot $i \in [N]$ and $k \in \{0, \ldots, K-1\}$. These factors are used to generate candidate trajectories for Cooperative Policy Selection (Sec.~\ref{sec: policy_selection}) and serve as the optimization variables in Temporal Adaptation (Sec.~\ref{sec: temporal_optimization}). The scaled learned \acp{ds} are therefore given by,
\begin{equation} \label{eq: ds_with_a}
\dot{\bm{x}}_{i}^{\text{d}}(k) =
\alpha_i(k) \bm{f}_{i,m_i}(\bm{x}_i(k)),
\quad i \in [N].
\end{equation}
The resulting end-effector velocity commands are mapped to joint velocities through the \ac{llc}. In this work, we employ differential null-space inverse kinematics~\citep{buss2004introductionIK},
\begin{equation}
    \dot{\bm{q}}_i^{\text{d}} = \bm{J}^\top \left( \bm{J}\bm{J}^\top + \bm{D} \right)^{-1} \dot{\bm{x}}_{i}^{\text{d}}
+
\left( \bm{I} - \bm{J}^\dagger \bm{J} \right)
\bm{K}_n \left( \bm{q}^{\text{H}} - \bm{q}_i \right),
\end{equation}
with Jacobian $\bm{J}(\bm{q}_i)$, damping matrix $\bm{D}$, null-space gain $\bm{K}_n$ and homing configuration $\bm{q}^{\text{H}}$ for the robot-arm joints only.
The resulting joint velocities are then integrated to obtain the next joint configuration. The discrete-time system can thus be expressed compactly as
\begin{equation} \label{eq: F_full_step}
\bm{q}_i(k+1) =
\bm{F}_{i, m_i}\big(\bm{q}_i(k),\alpha_i(k)\big),
\end{equation}
where $\bm{F}_{i, m_i}$ represents the combined effect of the \ac{fk}, the selected learned \ac{ds}, the LLC, and the discrete-time integration, with current robot configuration $\bm{q}_i(0) = \bm{q}_{i}^{0}$. The result is a simulated trajectory along the discrete horizon, 
$\bm{\tau}_{i,m_i} =
\left\{
\left(\bm{q}_i(k), \dot{\bm{q}}_i(k), \alpha_i(k)\right)
\right\}_{k=0}^{K},
\ \forall i \in [N].$

\textbf{Modulation.} 
Although static obstacle avoidance may be reflected in the demonstrations, it is not explicitly guaranteed by the learned \acp{ds}, which may generate trajectories intersecting static objects such as tables or cabinets. 
While our primary focus is on inter-robot collision avoidance, we additionally modulate each learned \ac{ds} in~\eqref{eq: ds_with_a} via $\bar{\bm{f}}(\bm{x}) = \bm{M}(\bm{x}) \bm{f}(\bm{x})$, with modulation matrix $\bm{M}(\bm{x})$~\citep{khansari2012modulation}. The modulation ellipse is fitted within the obstacle footprint while maintaining clearance from the demonstrated trajectory, enabling local deformation around the obstacle.

\subsection{Cooperative Policy Selection} \label{sec: policy_selection}
Cooperative Policy Selection evaluates and ranks candidate combinations of task strategies to identify policies suitable for collaborative execution, which are passed to the Temporal Adaptation planner.
For each combination $(m_i,m_j)$ of learned single-agent \acp{ds}, we perform forward simulations using~\eqref{eq: F_full_step} with $n_{\text{init}}$ parallel, predefined velocity-scaling configurations $\bm{\alpha} := [\bm{\alpha}_{i}, \bm{\alpha}_{j}]$. These include the nominal configuration $\bm{\alpha} = [\bm{1}, \bm{1}]$ and $(n_{\text{init}}-1)$ additional configurations with varying velocity-scaling factors along the trajectory.
Among these $n_{\text{init}}$ forward simulations, we retain the one with the lowest cost for each combination $(m_i,m_j)$.
First, combinations whose demonstrated target poses collide are discarded by assigning a high static cost $w_g$. The remaining combinations are ranked based on their demonstration closeness cost $J_{\text{demo}}$ and trajectory cost $J_{\tau}$,
\begin{equation}
J^{\text{rank}}_{(m_i, m_j)} =
\begin{cases}
w_{\text{g}}, & \text{if any } z \in Z_{\text{demos}}:
d(
\bm{x}_{m_i}^{\text{demo},z}(1),
\bm{x}_{m_j}^{\text{demo},z}(1)
) < d_{\min},\\
w_{\text{s}} J_{\text{demo}} + w_{\tau} J_{\tau}, & \text{otherwise},
\end{cases}
\end{equation}
with weights $w_{\text{s}}$ and $w_{\tau}$. 
The cost $J_{\text{demo}}$ averages the distances from each robot's initial pose to its closest point on the demonstration,
$
J_{\text{demo}} = \frac{1}{N}\sum_{i=1}^{N}\min_{z} d\left(\bm{x}(0), \tau^{\text{demo},z}_{m_i}\right).
$
The trajectory cost $J_{\tau}$ penalizes deviations from the learned \acp{ds} velocity commands, as well as violations of collision-avoidance, task order, and speed constraints, as detailed in Sec.~\ref{sec: temporal_optimization}.

Once the ranking of the combinations is established, the combination $(m_i, m_j)$ with the lowest cost is send to the Temporal Adaptation layer, which receives a computational budget for resolving the trajectories of all robots to meet the criteria.
A solution is accepted, if it meets the hard constraints for collision-avoidance, task-ordering and goal convergence. 
If the criteria is not met after optimization, temporal adaptation is performed of the next-best option in the ranking. If the criteria are met before optimization, the initial solution is accepted. 

\subsection{Temporal Adaptation} \label{sec: temporal_optimization}
Temporal Adaptation adjusts the velocity scaling factors of the learned \acp{ds} at each step along the horizon to minimize deviations from the selected policy combination $(m_i,m_j)$ while satisfying the required constraints. Unlike spatial collision avoidance, this approach preserves the motion direction prescribed by the learned vector fields, modifying only the velocity magnitude.
This enables the robots to coordinate their timing while retaining the learned motion patterns, resulting in predictable trajectories with minimal deviations from the intended motion.
We first formulate the collision-avoidance and task-ordering constraints, followed by the complete optimization problem.

\noindent \textbf{Robot-robot collision avoidance.}
To ensure whole-body collision avoidance, the distance between every pair of points on the robots' bodies must exceed a threshold $d_{\text{min}}$ and at every step $k \in \{1, \ldots, K\}$. This is expressed as,
\begin{equation} \label{eq: coll_avoid}
    d_{\text{min}} - d 
    \left(f_{\text{coll}}(\bm{q}_i(k)), f_{\text{coll}}(\bm{q}_j(k)) \right) \leq 0, \quad \forall i \neq j,
\end{equation}
where $f_{\text{coll}}$ maps each robot configuration to a geometric representation of its body, composed of primitives such as spheres or capsules. 
The constraint is evaluated for all pairs of primitives across all robots. For capsule-based representations, the distance is computed as the minimum distance between their centerline segments, minus the sum of their radii~\citep{ericson2004capsules}. These geometric representations provide compact, almost-everywhere differentiable distance functions suitable for gradient-based optimization.

\noindent \textbf{Task Order.} 
Some tasks require one robot to complete part of its task before the other can proceed, as in~\eqref{eq: task_order_problem_def}. For example, in the restaurant scene (Fig.~\ref{fig:overview}), a glass may only be placed after the table has been wiped underneath it. Let $k^{\text{c}}_1$ denote the time step at which the first robot completes the order-relevant part of the task, defined as the moment its end-effector $\bm{x}_1$ passes over the second robot's goal $\bm{x}_{2}^{\text{g}}$. The second robot must therefore remain outside a radius $r_{\mathrm{place}}$ of its goal until $k^{\text{c}}_1$. The resulting task-ordering constraint is defined as follows,
\begin{equation} \label{eq: task_order}
  \lVert \bm{x}_{2}(k) - \bm{x}_{2, \text{g}}\rVert \geq r_{\mathrm{place}}, \qquad \forall\, k < k^{\text{c}}_1(\bm{\alpha}_1, \bm{q}_1^0),
\end{equation}
where $k_1^{\text{c}}$ depends on $\bm{\alpha}_1$ and the initial configuration $\bm{q}_1^0$.

\noindent \textbf{Temporal optimization.}
For a system of $N$ robots, the resulting optimization problem determines the velocity scaling variables $\bm{\alpha}$ over the horizon, minimizing deviations from the learned dynamics while satisfying the collision-avoidance and task-ordering constraints,
\begin{subequations} \label{eq: all_optimization}
\begin{align}
    \min_{\bm{\alpha}} \ \ & \sum_i^N\left(w_\alpha \sum_{0}^{K-1} (\alpha_i(k) - 1)^2 \right) & \\ 
    s.t. \ \ & 
    \bm{q}_{i+1} = \bm{F}_{i, m_i}\big(\bm{q}_i(k),\alpha_i(k)\big), & \forall i \in [ N], \forall k \in \{0, \cdots, K -1\}\label{eq: 1_full_model}\\
    & \text{Eq.}~\eqref{eq: coll_avoid}, &\quad \forall k \in \{1, \cdots, K\} \label{eq: 1_coll_avoid}\\
    & \text{Eq.}~\eqref{eq: task_order}, &\quad \forall k \in \{1, \cdots, K\} \label{eq: 1_task_constraints}\\
        & |\dot{\bm{x}}_{i}^{\text{d}}| \leq \bar{v}_{\text{max}},  &\quad \forall i \in [N], \forall k \in \{0, \cdots, K-1\} \label{eq: 1_vmax}\\
    & 0 < \alpha_i(k) \leq \alpha_{\text{max}}, &\quad \forall i \in [N], \forall k \in \{0, \cdots, K-1\} \label{eq: 1_alpha}
\end{align}
\end{subequations}
The objective function penalizes deviations from nominal execution, $\alpha_i(k)=1$, for all robots and time steps, using weight $w_\alpha$.
The system dynamics in~\eqref{eq: 1_full_model} forward-simulate each robot's motion over the horizon leveraging $\bm{\alpha}$ and the learned single-agent \acp{ds}. 
Collision avoidance is enforced through~\eqref{eq: 1_coll_avoid}, whereas~\eqref{eq: 1_task_constraints} represents one possible task-ordering constraint.
The factors $\bm{\alpha}$ are constrained to be strictly positive and bounded to prevent backward motion and excessive scaling. Additionally, the desired end-effector velocity is limited by $\bar{v}_{\max}$.  
Constraints~\eqref{eq: 1_coll_avoid}--\eqref{eq: 1_vmax} are incorporated as penalty terms, while constraint~\eqref{eq: 1_alpha} is enforced through sigmoid reparameterization. The resulting optimization problem is solved using gradient-based methods in JAX, with gradients propagated through the differentiable learned \acp{ds}, which can be represented by \acp{dnn}.

\noindent \textbf{Notes on stability.} In continuous time, temporal scaling with $\alpha(t)>0$ reparameterizes time without changing the spatial trajectories of the learned \acp{ds}. Consequently, their asymptotic convergence properties are preserved.
However, the stability of the complete robot system in~\eqref{eq: 1_full_model} also depends on the low-level controller, robot kinematics, and constraints, which may prevent reaching the goal.
In discrete time, large scaling factors may cause numerical errors or overshooting, for which stabilizing bounds can be derived for certain \acp{ds}~\citep{lee2025behavior}. The bounds in~\eqref{eq: 1_vmax} and~\eqref{eq: 1_alpha}, on the desired end-effector velocity and $\alpha(k)$ respectively, help mitigate these effects and support stable execution.
By restricting adaptations to temporal scaling and explicitly enforcing goal-convergence and safety criteria during optimization, our approach aims to preserve the desirable properties of the learned behaviors while enabling safe collaborative execution.

\section{Experimental Results} \label{sec:experimental_results}
We evaluate the proposed framework in three cooperative mobile manipulation scenarios, describing the experimental setup and quantitatively analyzing the results. The supplementary material comprises an Appendix with additional details, and accompanying video (\href{https://youtu.be/9AiIJNsJPlQ}{https://youtu.be/9AiIJNsJPlQ}).

\subsection{Experimental Setup}
\noindent\textbf{Simulation Setup.} Experiments are conducted in simulation using two mobile manipulators, each consisting of a Clearpath Dingo omnidirectional base and a Kinova Gen3 Lite arm, resulting in a 9-DoF system. The \ac{fk} and \ac{llc} are parallelized using JAX~\citep{jaxopt_implicit_diff} and FRAX~\citep{morton2026frax}, a tailored version of MuJoCo XLA (MJX)~\citep{todorov2012mujoco} for robot kinematics. 
Each robot is approximated by 8 capsules, slightly over-approximating its geometry while enabling efficient collision-avoidance computations.

\begin{figure}
    \centering
    \includegraphics[width=\linewidth]{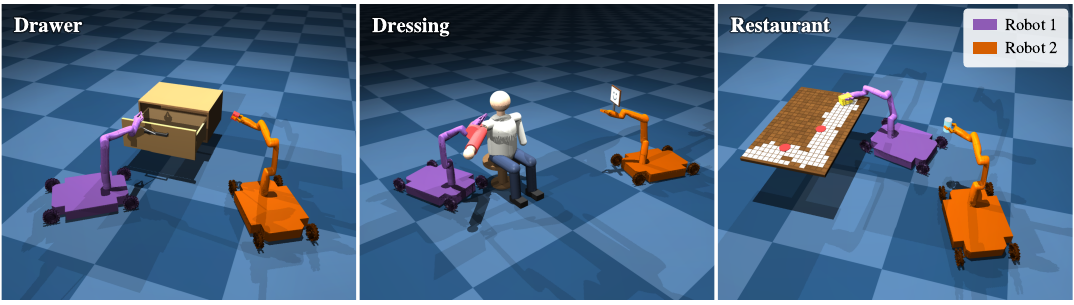}
    \caption{The three scenarios considered: Drawer, Dressing and Restaurant. The visuals are snapshots of successful runs of \acs{calds}-\acs{dsmpman}.}
    \label{fig:scenes}
\end{figure}

\noindent\textbf{Scenarios.}  We consider three cooperative scenarios (Fig.~\ref{fig:scenes}): (1) \emph{Restaurant}: Robot 1 (R1) cleans a table while Robot 2 (R2) places a cup on the cleaned portion, (2) \emph{Dressing}: R1 dresses a human arm while R2 uses a mirror to show different viewpoints, and (3) \emph{Drawer}: R1 opens a drawer while R2 places an item inside. To focus on cooperative motion adaptation, we simplify the simulated tasks: Dressing involves moving a cylinder over the human arm, while the drawer is attached to R1 by a flexible string, avoiding rigid grasp interactions. In Restaurant, R2 may place the cup only once R1 has cleaned the intended placement area. For mirror-showing, table-cleaning, and drawer-opening, the corresponding robot must first approach the task's starting region within a margin of $d_{\mathrm{start}}$, preventing the robot from starting the task midway.

\noindent\textbf{Learned \acp{ds} and Comparison.} We consider two methods for learning the single-agent DSs used by \ac{calds}: \ac{dsmpman}~\citep{bakker2026curve} and \ac{puma}~\citep{perez2024puma}. \ac{dsmpman} fits a curve on the manifold to demonstrated trajectories and defines a DS that attracts towards the curve while progressing along its direction. \ac{puma} instead learns a \ac{ds} using a \ac{dnn}, enforcing convergence through its loss function and connecting latent-space dynamics to task-space dynamics. Both methods support pose representations on $SE(3)$, guarantee convergence to the goal under their stated assumptinos, and can be learned from few demonstrations (1–10).

We compare against \ac{cdmp}~\citep{singh2025collaborative}, using its open-source implementation. We retain its collaborative adaptation framework but replace its skill library and policy selection with the same learned \acp{ds} and selection method used by \ac{calds}. As \ac{cdmp} was designed for dual manipulators, we adapt its implementation to mobile manipulators (Appendix~\ref{appendix: exp_details}). As a whole-body spatial-avoidance baseline, we replace temporal adaptation with a centralized \ac{cbf} safety filter~\citep{wang2017safety}, applied at the joint-velocity level~\citep{singletary2021safety}. Both robots execute the same learned \ac{ds} options selected by our ranking, with nominal timing. We refer to these baselines as \ac{cbf}-\ac{dsmpman} and \ac{cbf}-\ac{puma}, respectively. At each control step, a single \ac{qp} minimally modifies their nominal joint velocities subject to collision-avoidance constraints for the 32 closest capsule pairs. Finally, we consider an ablation using nominal learned DSs ($\bm{\alpha}=[\bm{1}, \bm{1}]$), with each robot's task strategy selected according to the closest goal to its initial end-effector position.

\noindent\textbf{Demonstrations and Experimental Parameters.} Demonstrations are collected through a teleoperation interface. For each task-option, $m_i \in [M_i]$, we collect five demonstrations: one illustrating the desired motion and four showing adaptations from different initial configurations. \ac{dsmpman} and \ac{cdmp} use the first demonstration to define the reference motion, whereas PUMA trains its DNN on all five to learn adaptations to different initial configurations.

Each robot uses a planning horizon of $K=300$ steps, corresponding to $30\unit{\second}$. Initial configurations are randomized, and further experimental details and parameters are provided in Appendix~\ref{appendix: exp_details}. For static obstacle avoidance (table, dresser, and chair), we fit an ellipse to modulate the learned DS, maintaining $0.1\unit{\meter}$ clearance from the first demonstration to preserve target reachability. We use the L-BFGS solver from JAXopt~\citep{blondel2022jaxopt}, with time budgets of $1\unit{\second}$ for \ac{calds}-\ac{dsmpman} and $3\unit{\second}$ for \ac{calds}-\ac{puma}, accounting for \ac{puma}'s approximately threefold longer rollout time to allow comparable solver iterations.

\noindent\textbf{Performance Metrics.} We evaluate \ac{calds}-\ac{dsmpman} and \ac{calds}-\ac{puma} against \ac{cbf}-\ac{dsmpman}, \ac{cbf}-\ac{puma}, \ac{cdmp} and the aforementioned ablation. A run is considered successful if it reaches the goal, is collision-free and meets task ordering constraints, if applicable to the scenario. \textit{Goal reaching} requires the final end-effector position to be within $0.15\unit{\meter}$ of the desired goal. \textit{Whole-body collision avoidance} evaluates if the minimum robot-robot capsule distance is positive and whether the end-effector governed by the \ac{ds} enters the static obstacle ellipse. 
For the Restaurant scenario, \textit{task ordering} requires the cup to be placed on the cleaned portion of the table.
In addition to the success rate, we report the \emph{Path Deviation}, defined as the maximum discrete Fréchet distance across robots, between their end-effector paths and the nominal paths ($\bm{\alpha}=[\bm{1}, \bm{1}]$).
In addition to the success rate, we report three computation times. \emph{Training/fitting time} captures the average offline computation required to obtain each learned \ac{ds}. \emph{Planning time} measures the time required to generate cooperative trajectories for both robots, including Cooperative Policy Selection and Temporal Adaptation. Finally, \emph{query time} measures the time required to compute a single action for both robots.

\subsection{Experimental Analysis}
\begin{figure}
    \centering
    \includegraphics[width=\linewidth]{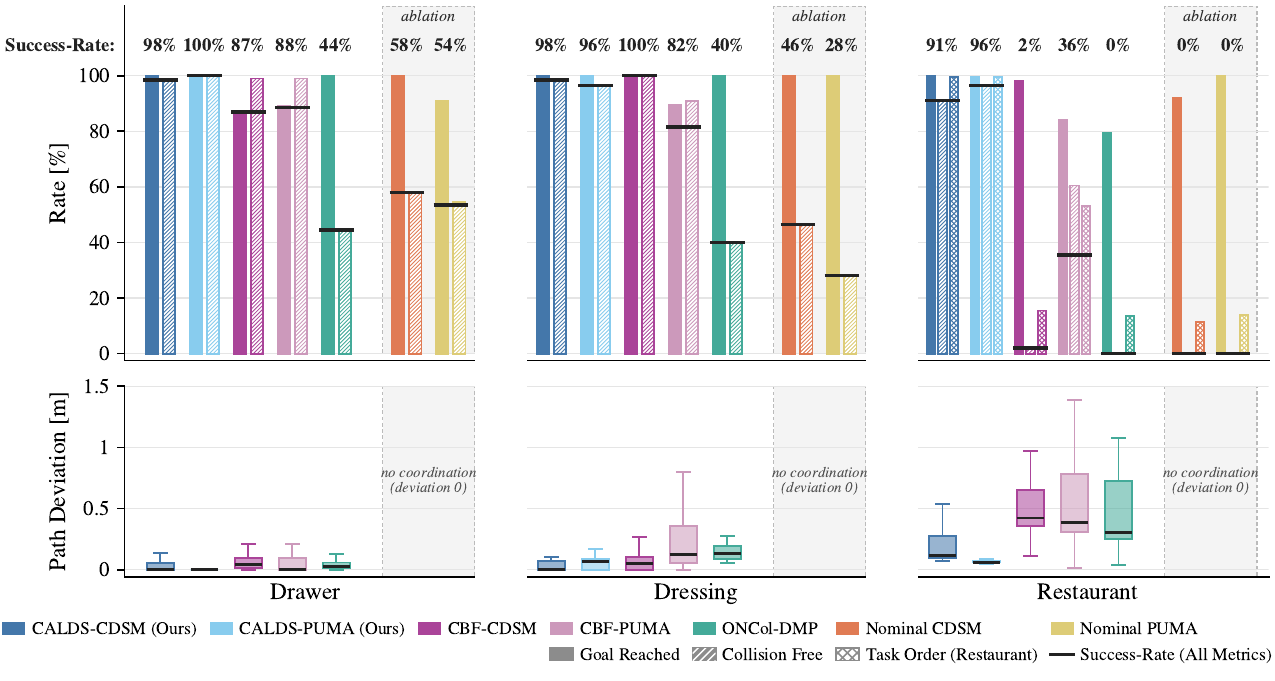}
    \caption{
    Results over 200 randomized runs per scenario, reporting the overall success rate, the individual metrics, i.e. goal reaching, collision avoidance, and task ordering (restaurant scenario), and path deviation.}
    \label{fig:success-rate}
\end{figure}
When analyzing Fig.~\ref{fig:success-rate}, we observe that the proposed \ac{calds}-\ac{dsmpman} and \ac{calds}-\ac{puma} enable cooperative adaptation of learned \acp{ds} in all three scenarios, considerably improving the success rate (91-100\%) considerably, compared to Nominal \ac{dsmpman} and Nominal \ac{puma}, while achieving small path deviations.
Further improvements could be achieved through longer planning horizons, more parallel runs with diverse initial guesses, increased optimization time, and allowing multiple fallbacks to lower-ranked policy combinations. However, these improvements would come at the expense of real-time adaptability.
An example of the resulting trajectories is shown in Fig.~\ref{fig:overview} and is further illustrated in the accompanying video.

Compared to \ac{cbf}-\ac{dsmpman} and \ac{cbf}-\ac{puma}, respectively, the proposed \ac{calds}-\ac{dsmpman} and \ac{calds}-\ac{puma} achieve smaller path deviations, indicating that temporal adaptation remains closer to the path of the nominal learned \acp{ds}.
Moreover, the success rates of the CBF-based approaches drop drastically in the Restaurant scenario, as their local collision-avoidance constraints can lead to conflicting requirements when avoiding both the static obstacle and the other robot, and task-ordering constraints are not enforced.
\ac{cdmp} achieves end-effector collision-free trajectories in 98.5$\%$ of the cases, comparable to the proposed \ac{calds}-\ac{dsmpman} and \ac{calds}-\ac{puma} (100$\%$ and 99.5$\%$, respectively). However, it fails to avoid whole-body collisions, highlighting the limitation of its temporal adaptation in guaranteeing whole-body collision avoidance, as also acknowledged by the authors~\citep{singh2025collaborative}. Moreover, task ordering is not explicitly enforced in the Restaurant scenario, even when R2 is instructed to wait for R1 when their temporal adaptation is triggered.

As shown in Table~\ref{table: computaton times}, fitting times for \ac{calds}-\ac{dsmpman} are only $1.3\pm0.5\unit{\second}$, representing a negligible overhead compared to the several minutes required for collecting demonstrations. Training \ac{calds}-\ac{puma} takes approximately $11\unit{\minute}$ per \ac{dnn}, which remains reasonable given the offline nature of this process.
The planning time for \ac{calds}-\ac{puma} ($1.49 \pm 1.73\unit{\second}$) is approximately twice that of \ac{calds}-\ac{dsmpman} ($0.76 \pm 0.64\unit{\second}$), reflecting both the higher computational cost of forward simulation in \ac{puma} and its larger planning budget of $3\unit{\second}$, compared to $1\unit{\second}$ for \ac{calds}-\ac{dsmpman}.
Nevertheless, both approaches achieve relatively low computation times considering whole-body collision avoidance over a $30\unit{\second}$ horizon, with individual action query times of $0.18\unit{\milli\second}$. For real-time adaptation in uncertain dynamic environments, however, a low-level planner capable of handling dynamic changes at \qtyrange{10}{100}{\hertz} may still be required, or further computational improvements to the proposed approach may be necessary.
Closed-loop simulation results for our approach and the baselines, incorporating per-time-step joint-configuration noise and subject to a strict planning time limit of $1\unit{\second}$, are provided in Appendix~\ref{appendix: closed_loop}.

\begin{table}
\centering
\begin{tabular}{lrrrrrrrr}
\toprule
 & \multicolumn{1}{c}{Training/Fitting time [s]} & \multicolumn{1}{c}{Planning time [s]} & \multicolumn{1}{c}{Query time [ms]} \\
\midrule
CALDS-CDSM (Ours)       & \phantom{00}1.3 $\pm$ \phantom{00}0.5     & 0.76 $\pm$ 0.64   &  0.18 $\pm$ 0.06 \\
CALDS-PUMA (Ours)       & 652.0 $\pm$ 234.2                         & 1.49 $\pm$ 1.73   &  0.19 $\pm$ 0.21 \\
CBF-CDSM                & \phantom{00}1.3 $\pm$ \phantom{00}0.5     & 0.45 $\pm$ 0.10   &  0.43 $\pm$ 0.14 \\
CBF-PUMA                & 652.0 $\pm$ 234.2                         & 0.67 $\pm$ 0.17   &  0.45 $\pm$ 0.23 \\
ONCol-DMP               & \phantom{000000}-                         & 0.80 $\pm$ 0.39   &  2.98 $\pm$ 0.04 \\
\bottomrule
\end{tabular}
\caption{Computational costs for the evaluated methods, reporting mean $\pm$ standard deviation of training/fitting time, planning time, and query time, averaged over 600 runs.
The planning time of \ac{cbf}-\ac{dsmpman} and \ac{cbf}-\ac{puma} is attributed to the cooperative policy selection procedure, whereas that of \ac{cdmp} includes both policy selection and DMP generation.
}
\label{table: computaton times}
\end{table}
\section{Conclusion and Future Work}
This work introduced \ac{calds}, a framework for cooperative adaptation of learned \acp{ds} for mobile manipulators, integrating cooperative policy selection and temporal adaptation. We demonstrated its applicability to collaborative tasks involving two mobile manipulators across three distinct scenarios. The resulting trajectories preserve the behavior encoded by the learned \acp{ds}, while satisfying whole-body collision avoidance and task requirements. Furthermore, exploiting the stability properties of the learned \acp{ds} enables learning from few demonstrations and supports pose representations on $SE(3)$.
Future work will investigate integrating the framework into a single neural network conditioned on discrete markers representing different behavioral modes~\citep{sun2021task}, as well as extending it to human-robot collaboration, where increased uncertainty and behavioral diversity introduce additional challenges.

\subsection*{Generative AI Use}
Generative AI tools were used to assist with code generation, text revision, and figure construction. All scenes and trajectories presented in the paper originate from simulated experiments and were not AI-generated. The authors have reviewed and verified the AI-assisted content and take full responsibility for the accuracy and integrity of the final manuscript and results.

\acks{This project has received funding from the European Union through ERC,
INTERACT, under Grant 101041863. Views and opinions expressed are however those of the author(s) only and do not necessarily reflect those of the European Union. Neither the European Union nor the granting authority
can be held responsible for them.}

\bibliography{references}

\begin{thebibliography}{33}
\providecommand{\natexlab}[1]{#1}
\providecommand{\url}[1]{\texttt{#1}}
\expandafter\ifx\csname urlstyle\endcsname\relax
  \providecommand{\doi}[1]{doi: #1}\else
  \providecommand{\doi}{doi: \begingroup \urlstyle{rm}\Url}\fi

\bibitem[Bakker et~al.(2026)Bakker, Schonger, L{\"o}w, Alonso-Mora, and Calinon]{bakker2026curve}
Saray Bakker, Martin Schonger, Tobias L{\"o}w, Javier Alonso-Mora, and Sylvain Calinon.
\newblock Curve-induced dynamical systems on {Riemannian} manifolds and {Lie} groups.
\newblock \emph{arXiv preprint arXiv:2603.05268}, 2026.

\bibitem[Billard et~al.(2022)Billard, Mirrazavi, and Figueroa]{billard2022learning}
Aude Billard, Sina Mirrazavi, and Nadia Figueroa.
\newblock \emph{Learning for adaptive and reactive robot control: a dynamical systems approach}.
\newblock {MIT} Press, 2022.

\bibitem[Blondel et~al.(2022{\natexlab{a}})Blondel, Berthet, Cuturi, Frostig, Hoyer, Llinares-L{\'o}pez, Pedregosa, and Vert]{blondel2022jaxopt}
Mathieu Blondel, Quentin Berthet, Marco Cuturi, Roy Frostig, Stephan Hoyer, Felipe Llinares-L{\'o}pez, Fabian Pedregosa, and Jean-Philippe Vert.
\newblock Efficient and modular implicit differentiation.
\newblock In \emph{Advances in Neural Information Processing Systems (NeurIPS)}, 2022{\natexlab{a}}.

\bibitem[Blondel et~al.(2022{\natexlab{b}})Blondel, Berthet, Cuturi, Frostig, Hoyer, Llinares-L{\'o}pez, et~al.]{jaxopt_implicit_diff}
Mathieu Blondel, Quentin Berthet, Marco Cuturi, Roy Frostig, Stephan Hoyer, Felipe Llinares-L{\'o}pez, et~al.
\newblock Efficient and modular implicit differentiation.
\newblock \emph{Advances in neural information processing systems}, 35:\penalty0 5230--5242, 2022{\natexlab{b}}.

\bibitem[Bradbury et~al.(2018)Bradbury, Frostig, Hawkins, Johnson, Leary, Maclaurin, et~al.]{jax2018github}
James Bradbury, Roy Frostig, Peter Hawkins, Matthew~James Johnson, Chris Leary, Dougal Maclaurin, et~al.
\newblock {JAX}: composable transformations of {P}ython+{N}um{P}y programs, 2018.
\newblock URL \url{http://github.com/jax-ml/jax}.

\bibitem[Buss et~al.(2004)]{buss2004introductionIK}
Samuel~R Buss et~al.
\newblock Introduction to inverse kinematics with jacobian transpose, pseudoinverse and damped least squares methods.
\newblock \emph{IEEE Journal of Robotics and Automation}, 17\penalty0 (1-19):\penalty0 16, 2004.

\bibitem[Dong et~al.(2026)Dong, Bhatt, Choi, and Mehr]{dong2025mimic}
Dayi Dong, Maulik Bhatt, Seoyeon Choi, and Negar Mehr.
\newblock Mimic-d: Multi-modal imitation for multi-agent coordination with decentralized diffusion policies.
\newblock In \emph{Proc.\ {IEEE} International Conference on Robotics and Automation ({ICRA})}, pages 17193--17200. IEEE, 2026.

\bibitem[Dong et~al.(2025)Dong, Zhang, Zhang, Ling, Fu, Bai, M{\'a}rton, Bing, Chen, Knoll, et~al.]{dong2025m4diffuser}
Ju~Dong, Lei Zhang, Liding Zhang, Yao Ling, Yu~Fu, Kaixin Bai, Zolt{\'a}n-Csaba M{\'a}rton, Zhenshan Bing, Zhaopeng Chen, Alois~Christian Knoll, et~al.
\newblock M4diffuser: Multi-view diffusion policy with manipulability-aware control for robust mobile manipulation.
\newblock \emph{arXiv preprint arXiv:2509.14980}, 2025.

\bibitem[Ericson(2005)]{ericson2004capsules}
Christer Ericson.
\newblock \emph{Real-Time Collision Detection}.
\newblock Morgan Kaufmann, San Francisco, 2005.

\bibitem[Fu et~al.(2024)Fu, Zhao, and Finn]{fu2024mobile}
Zipeng Fu, Tony~Z. Zhao, and Chelsea Finn.
\newblock Mobile aloha: Learning bimanual mobile manipulation with low-cost whole-body teleoperation.
\newblock In \emph{Proc.\ Conference on Robot Learning ({CoRL})}, 2024.

\bibitem[Hartmann et~al.(2022)Hartmann, Orthey, Driess, Oguz, and Toussaint]{hartmann2022long}
Valentin~N Hartmann, Andreas Orthey, Danny Driess, Ozgur~S Oguz, and Marc Toussaint.
\newblock Long-horizon multi-robot rearrangement planning for construction assembly.
\newblock \emph{IEEE Transactions on Robotics}, 39\penalty0 (1):\penalty0 239--252, 2022.

\bibitem[He et~al.(2025)He, Camps, Liu, Schwager, and Sartoretti]{he2025latent}
Chengyang He, Gadiel~Sznaier Camps, Xu~Liu, Mac Schwager, and Guillaume Sartoretti.
\newblock Latent theory of mind: A decentralized diffusion architecture for cooperative manipulation.
\newblock \emph{Proc.\ Conference on Robot Learning ({CoRL})}, 2025.

\bibitem[Khansari-Zadeh and Billard(2012)]{khansari2012modulation}
Seyed~Mohammad Khansari-Zadeh and Aude Billard.
\newblock A dynamical system approach to realtime obstacle avoidance.
\newblock \emph{Autonomous Robots}, 32\penalty0 (4):\penalty0 433--454, 2012.

\bibitem[Lee et~al.(2025)Lee, Lee, Ha, and Park]{lee2025behavior}
Byeongho Lee, Yonghyeon Lee, Junsu Ha, and Frank~C Park.
\newblock Behavior-controllable stable dynamics models on {Riemannian} configuration manifolds.
\newblock \emph{IEEE Transactions on Robotics}, 2025.

\bibitem[Morton and Pavone(2026)]{morton2026frax}
Daniel Morton and Marco Pavone.
\newblock frax: Fast robot kinematics and dynamics in jax.
\newblock \emph{arXiv preprint arXiv:2604.04310}, 2026.

\bibitem[P{\'e}rez-Dattari et~al.(2024)P{\'e}rez-Dattari, Della~Santina, and Kober]{perez2024puma}
Rodrigo P{\'e}rez-Dattari, Cosimo Della~Santina, and Jens Kober.
\newblock Puma: Deep metric imitation learning for stable motion primitives.
\newblock \emph{Advanced Intelligent Systems}, 6\penalty0 (11):\penalty0 2400144, 2024.

\bibitem[P{\'e}rez-Dattari et~al.(2026)P{\'e}rez-Dattari, Leiva, Testa, Rozo, del Solar, and Jaquier]{perez2026letthedynamicsflow}
Rodrigo P{\'e}rez-Dattari, Francisco Leiva, Andrea Testa, Leonel Rozo, Javier~Ruiz del Solar, and No{\'e}mie Jaquier.
\newblock Let the dynamics flow: Stable flow matching dynamical systems.
\newblock \emph{arXiv preprint arXiv:2606.03834}, 2026.

\bibitem[Peters et~al.(2026)Peters, Ferranti, Bajcsy, and Alonso-Mora]{peters2026coordinated}
Lasse Peters, Laura Ferranti, Andrea Bajcsy, and Javier Alonso-Mora.
\newblock Coordinated diffusion: Generating multi-agent behavior without multi-agent demonstrations.
\newblock \emph{arXiv preprint arXiv:2605.11485}, 2026.

\bibitem[Ratliff et~al.(2018)Ratliff, Issac, Kappler, Birchfield, and Fox]{ratliff_riemannian_2018}
Nathan~D Ratliff, Jan Issac, Daniel Kappler, Stan Birchfield, and Dieter Fox.
\newblock Riemannian motion policies.
\newblock \emph{arXiv preprint arXiv:1801.02854}, 2018.

\bibitem[Reuss et~al.(2023)Reuss, Li, Jia, and Lioutikov]{reuss2023goal}
Moritz Reuss, Maximilian Li, Xiaogang Jia, and Rudolf Lioutikov.
\newblock Goal-conditioned imitation learning using score-based diffusion policies.
\newblock \emph{Proc. of Robotics: Science and Systems}, 2023.

\bibitem[Rudy et~al.(2022)Rudy, Idzikowski, Roszkowska, and Kluwak]{rudy2022multiple}
Jaros{\l}aw Rudy, Rados{\l}aw Idzikowski, Elzbieta Roszkowska, and Konrad Kluwak.
\newblock Multiple mobile robots coordination in shared workspace for task makespan minimization.
\newblock \emph{Processes}, 10\penalty0 (10):\penalty0 2087, 2022.

\bibitem[Saha and Julius(2017)]{saha2017task}
Sayan Saha and Anak~Agung Julius.
\newblock Task and motion planning for manipulator arms with metric temporal logic specifications.
\newblock \emph{{IEEE} Robotics and Automation Letters}, 3\penalty0 (1):\penalty0 379--386, 2017.

\bibitem[S{\'a}nchez-Molina et~al.(2024)S{\'a}nchez-Molina, Rodr{\'\i}guez, Moreno, S{\'a}nchez-Hermosilla, and Gim{\'e}nez]{sanchez2024greenhouses}
Jorge~Antonio S{\'a}nchez-Molina, Francisco Rodr{\'\i}guez, Jos{\'e}~Carlos Moreno, Juli{\'a}n S{\'a}nchez-Hermosilla, and Antonio Gim{\'e}nez.
\newblock Robotics in greenhouses. scoping review.
\newblock \emph{Computers and Electronics in Agriculture}, 219:\penalty0 108750, 2024.

\bibitem[Singh et~al.(2025)Singh, Xu, and Chang]{singh2025collaborative}
Siddharth Singh, Tian Xu, and Qing Chang.
\newblock Collaborative motion planning for multi-manipulator systems through reinforcement learning and dynamic movement primitives.
\newblock In \emph{Proc.\ {IEEE} International Conference on Robotics and Automation ({ICRA})}, pages 3369--3375, 2025.

\bibitem[Singletary et~al.(2021)Singletary, Kolathaya, and Ames]{singletary2021safety}
Andrew Singletary, Shishir Kolathaya, and Aaron~D Ames.
\newblock Safety-critical kinematic control of robotic systems.
\newblock \emph{IEEE Control Systems Letters}, 6:\penalty0 139--144, 2021.

\bibitem[Sun et~al.(2021)Sun, Probst, Paudel, Popovi{\'c}, Kanakis, Patel, Dai, and Van~Gool]{sun2021task}
Guolei Sun, Thomas Probst, Danda~Pani Paudel, Nikola Popovi{\'c}, Menelaos Kanakis, Jagruti Patel, Dengxin Dai, and Luc Van~Gool.
\newblock Task switching network for multi-task learning.
\newblock In \emph{2021 IEEE/CVF International Conference on Computer Vision (ICCV)}, pages 8271--8280. IEEE, 2021.

\bibitem[Sundaresan et~al.(2025)Sundaresan, Malhotra, Miao, Yang, Wu, Hu, Antonova, Engelmann, Sadigh, and Bohg]{sundaresan2025homer}
Priya Sundaresan, Rhea Malhotra, Phillip Miao, Jingyun Yang, Jimmy Wu, Hengyuan Hu, Rika Antonova, Francis Engelmann, Dorsa Sadigh, and Jeannette Bohg.
\newblock Homer: Learning in-the-wild mobile manipulation via hybrid imitation and whole-body control.
\newblock \emph{arXiv preprint arXiv:2506.01185}, 2025.

\bibitem[Todorov et~al.(2012)Todorov, Erez, and Tassa]{todorov2012mujoco}
Emanuel Todorov, Tom Erez, and Yuval Tassa.
\newblock Mujoco: A physics engine for model-based control.
\newblock \emph{Proc.\ {IEEE/RSJ} International Conference on Intelligent Robots and Systems ({IROS})}, 2012.

\bibitem[Tracy and Manchester(2024)]{tracy2024qpax}
Kevin Tracy and Zachary Manchester.
\newblock On the differentiability of the primal-dual interior-point method.
\newblock \emph{arXiv preprint arXiv:2406.11749}, 2024.

\bibitem[Urain et~al.(2022)Urain, Tateo, and Peters]{urain2022learning}
Julen Urain, Davide Tateo, and Jan Peters.
\newblock Learning stable vector fields on {Lie} groups.
\newblock \emph{IEEE Robotics and Automation Letters}, 7\penalty0 (4):\penalty0 12569--12576, 2022.

\bibitem[Wang et~al.(2017)Wang, Ames, and Egerstedt]{wang2017safety}
Li~Wang, Aaron~D Ames, and Magnus Egerstedt.
\newblock Safety barrier certificates for collisions-free multirobot systems.
\newblock \emph{IEEE Transactions on Robotics}, 33\penalty0 (3):\penalty0 661--674, 2017.

\bibitem[Wu et~al.(2023)Wu, Antonova, Kan, Lepert, Zeng, Song, Bohg, Rusinkiewicz, and Funkhouser]{wu2023tidybot}
Jimmy Wu, Rika Antonova, Adam Kan, Marion Lepert, Andy Zeng, Shuran Song, Jeannette Bohg, Szymon Rusinkiewicz, and Thomas Funkhouser.
\newblock Tidybot: Personalized robot assistance with large language models.
\newblock \emph{Autonomous Robots}, 47\penalty0 (8):\penalty0 1087--1102, 2023.

\bibitem[Zhao et~al.(2023)Zhao, Kumar, Levine, and Finn]{zhao2023learning}
Tony~Z Zhao, Vikash Kumar, Sergey Levine, and Chelsea Finn.
\newblock Learning fine-grained bimanual manipulation with low-cost hardware.
\newblock \emph{Proc. of Robotics: Science and Systems}, 2023.

\end{thebibliography}

\appendix
\section{Experimental Details} \label{appendix: exp_details}
Table~\ref{tab:parameters} summarizes the experimental setup and parameters used in the simulations, including randomized initial conditions, simulation settings, method-specific parameters for PUMA, CDSM and the baselines, temporal adaptation settings, and dynamical-system parameters. Upon publication, we will open-source our code, including examples and the evaluation of the three scenarios.

\textbf{\acs{cdmp}}: As \acs{cdmp} was originally designed for dual manipulators, we adapt the robot model to our mobile manipulators and modify the low-level controller to apply homing only to the manipulator, consistent with our approach. Finally, we increase the collision-checking distance threshold in the temporal adaptation from $0.5\unit{\meter}$ to $1.0\unit{\meter}$ (Tab.~\ref{tab:parameters}) to account for the larger robot size.

\begin{table}[t]
    \centering
    \footnotesize
    \begin{threeparttable}
    \caption{Experimental parameters.}
    \label{tab:parameters}
    \setlength{\tabcolsep}{5pt}
    \begin{tabular}{@{}lclc@{}}
        \toprule
        \textbf{Parameter} & \textbf{Value} &
        \textbf{Parameter} & \textbf{Value} \\
        \midrule
        \multicolumn{4}{@{}l}{\emph{Randomized start}\tnote{a}} \\
        Base-position perturbation, $\pm r_{\max}$~[\unit{\meter}] & $\pm 2.0$ &
        Min.\ robot--robot distance [\unit{\meter}] & 0.1 \\
        Base heading perturbation, $\pm\Delta\theta_{\max}$ [\unit{\degree}] & $\pm 30$ &
        Min.\ distance to static objects [\unit{\meter}] & 1.0 \\
        Arm joint perturbation, $\pm\Delta q_{\max}$ [\unit{\degree}] & $\pm 5$ &
        Runs per scenario & 200 \\
        \midrule
        \multicolumn{4}{@{}l}{\emph{Simulation}} \\
        Time step $\Delta t$ [\unit{\second}] & 0.1 &
        Horizon $N$ & 300 (30\unit{\second}) \\
        \midrule
        \multicolumn{4}{@{}l}{\emph{PUMA}} \\
        Optimizer & Adam &
        Learning rate & $10^{-4}$ \\
        Iterations (training) & 8\,000 &
        Activation function & GELU \\
        Number of layers ($\phi_\theta,\rho_\theta$) & (3, 3) &
        Latent-space dimension & 300 \\
        \midrule
        \multicolumn{4}{@{}l}{\emph{CDSM}} \\
        Curve segments & 8 &
        Basis functions & 3 \\
        Normal/Nominal twist factor & 7.0/3.0 &
        Max.\ iterations (fitting) & 200 \\
        \midrule
        \multicolumn{4}{@{}l}{\emph{Temporal adaptation}} \\
        Range of $\alpha$ & $[0,3]$  & Collision margin\tnote{b} \ ~[\unit{\meter}] & 0.2\\
        Initializations ($n_{\text{init}}$) & 4 &
        Solver & L-BFGS (5 iter.)\\
        Time budget [\unit{\second}] & 1 (CDSM), 3 (PUMA) & \\
        \midrule
        \multicolumn{4}{@{}l}{\emph{CBF}} \\
        Capsules-pairs per step $k$ & 32 &
        Collision margin\tnote{b} \ ~[\unit{\meter}] & 0.2 \\
        QP solver & qpax\tnote{c} & & \\
        \midrule
        \multicolumn{4}{@{}l}{\emph{\acs{cdmp}}} \\
        Collision distance (temporal)~[\unit{\meter}] & 1.0 &
        Robot that slows down (temporal) & R2 \\
        Collision margin (end-effector)~[\unit{\meter}] & 0.15 & & \\
        \midrule
        \multicolumn{4}{@{}l}{\emph{Dynamical systems}} \\
        Start distance $d_{\text{start}}$ [\unit{\meter}] & 0.1 & Ellipse clearance [\unit{\meter}] & 0.1\\
        First attract to start region (R1/R2) & \multicolumn{3}{c}{
        Drawer: $\checkmark$/--,\quad
        Dressing: $\checkmark$/$\checkmark$,\quad
        Restaurant: $\checkmark$/--} \phantom{00000000000}\\
        \bottomrule
    \end{tabular}
    \begin{tablenotes}[flushleft]
        \footnotesize
        \item[a] For each trial, the start configuration of a randomly selected demonstration is taken. The base position of each robot is then displaced by an offset sampled uniformly within a disc of radius $r_{\max}$, and its heading and arm joint angles are perturbed by offsets sampled uniformly within $\pm\Delta\theta_{\max}$ and $\pm\Delta q_{\max}$, respectively. Samples where a robot is too close to obstacles or to the other robot are rejected and redrawn.
        \item[b] Only used within temporal optimization and QP (CBFs); collision checking uses a margin of 0.0~m.
        \item[c] qpax is developed in~\citep{tracy2024qpax}.
    \end{tablenotes}
    \end{threeparttable}
\end{table}

\section{Closed-Loop Evaluation under State Noise} \label{appendix: closed_loop}
As a complementary evaluation, we investigate the robustness of Temporal Adaptation under state noise during closed-loop execution in the same three scenarios as Sec.~5. The planned motions are executed in MuJoCo, with Gaussian noise added to the state estimates at every control step. The noise standard deviations are $0.005\unit{\meter}$ for the base position and $0.005 \unit{\radian}$ for the base orientation and manipulator joint positions. 
Cooperative Policy Selection is performed only once at the beginning of each run, to isolate the effect of state estimation noise on Temporal Adaptation. 

The control loop runs at $\Delta t = 0.1\unit{\second}$ ($10~\unit{\hertz}$), while Temporal Adaptation replans at $1\unit{\hertz}$, allowing ten control steps between updates. To reflect a limited computational budget, the planning time is capped at $1\unit{\second}$ per update. In contrast, \acs{cdmp}, \acs{cbf}-\acs{puma}, and \acs{cbf}-\acs{dsmpman} replan at every control step ($10\unit{\hertz}$), as they support faster updates.
At each replanning step of the Temporal Adaptation, the velocity scaling factors $\bm{\alpha}$ from the previous iteration are shifted in time to align with the current timestep and used to warm-start the optimization. If the resulting trajectory maintains a minimum robot--robot distance above the collision margin of $0.2\unit{\meter}$ (Tab.~\ref{tab:parameters}), it is accepted without further optimization. Otherwise, Temporal Adaptation is triggered to compute an updated feasible solution within the available planning time.

\begin{figure}[h]
    \centering
    \includegraphics[width=\linewidth]{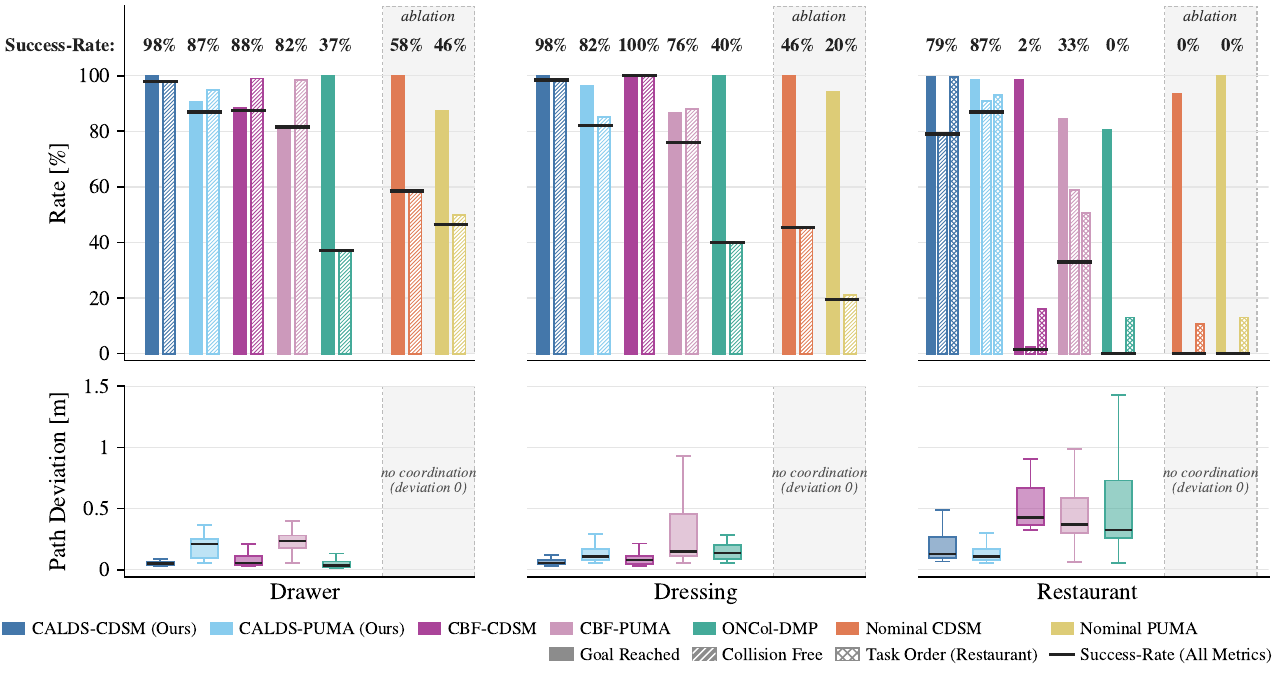}
    \caption{Experimental results of closed-loop simulations under state noise. Results over 200 randomized runs per scenario, reporting the overall success rate, the individual metrics, i.e. goal reaching, collision avoidance, and task ordering (restaurant scenario), and path deviation.}
    \label{fig:closed_loop}
\end{figure}

Figure~\ref{fig:closed_loop} shows that \acs{calds}-\acs{dsmpman} preserves its high success-rate in the Drawer and Dressing scenarios (98\%, 98\%), with a reduction only in the Restaurant scenario due to collisions (79\% compared to 91\% in Fig.~3). \acs{calds}-\acs{puma} experiences a reduced success-rate (87\%, 82\%, 87\%) across all three scenarios compared to Fig.~3 (100\%, 96\%, 96\%). This degradation likely stems from \acs{puma} requiring approximately three times longer to perform a forward simulation than \acs{dsmpman}, making it more affected by the $1\unit{\second}$ update cap. In addition, \acs{puma} attracts trajectories towards the goal rather than the demonstrated region, making it more susceptible to deviations from the nominal path under state noise.  
Despite operating at a substantially lower replanning frequency than the baselines, our method maintains effective cooperative multi-robot motion across all three scenarios under state noise, while accomplishing small path deviations.

\end{document}